\def\year{2018}\relax
\documentclass[letterpaper]{article} 
\usepackage{aaai18}  
\usepackage{times}  
\usepackage{helvet}  
\usepackage{courier}  
\usepackage{url}  
\usepackage{graphicx}  
\usepackage{amsmath}
\begin{document}
%
\title{Multi-Scale Face Restoration with Sequential Gating Ensemble Network}
\author{Jianxin Lin, Tiankuang Zhou, Zhibo Chen\\
University of Science and Technology of China, Hefei, China\\
\{linjx, zhoutk\}@mail.ustc.edu.cn, chenzhibo@ustc.edu.cn\\}
\maketitle
\begin{abstract}
Restoring face images from distortions is important in face recognition applications and is challenged by multiple scale issues, which is still not well-solved in research area. In this paper, we present a Sequential Gating Ensemble Network (SGEN) for multi-scale face restoration issue. We first employ the principle of ensemble learning into SGEN architecture design to reinforce predictive performance of the network. The SGEN aggregates multi-level base-encoders and base-decoders into the network, which enables the network to contain multiple scales of receptive field. Instead of combining these base-en/decoders directly with non-sequential operations, the SGEN takes base-en/decoders from different levels as sequential data. Specifically, the SGEN learns to sequentially extract high level information from base-encoders in bottom-up manner and restore low level information from base-decoders in top-down manner. Besides, we propose to realize bottom-up and top-down information combination and selection with Sequential Gating Unit (SGU). The SGU sequentially takes two inputs from  different levels and decides the output based on one active input. Experiment results demonstrate that our SGEN is more effective at multi-scale human face restoration with more image details and less noise than state-of-the-art image restoration models. By using adversarial training, SGEN also produces more visually preferred results than other models through subjective evaluation.
\end{abstract}

\section{Introduction}\label{intro}
\begin{figure}
  \centerline{\includegraphics[width=8.5cm]{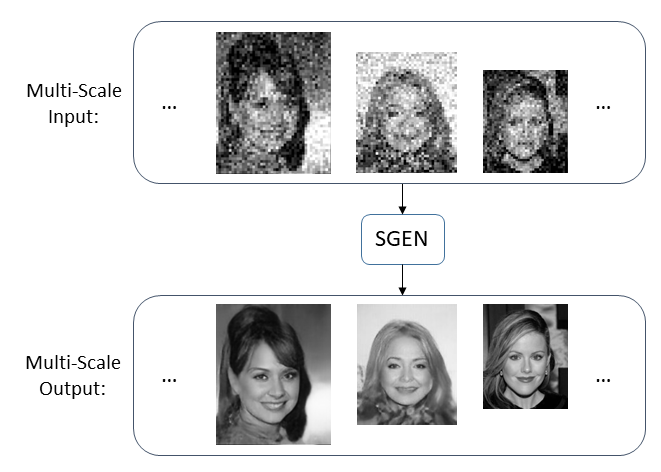}}
  \caption{Illustration of our SGEN on multi-scale face restoration. The multi-scale noise corrupted LR face images are up-sampled to the size of ground truth before being fed into network.}
  \centering
\label{fig:fig3}
\end{figure}
In the past decades, facial analysis techniques, such as face recognition and face detection, have achieved great progress. Meanwhile, thanks to the rapid development of surveillance system, facial analysis techniques have been employed to various applications, such as criminal investigation. However, the performance of most facial analysis techniques may degrade rapidly when given low quality face images. In real surveillance systems, the quality of the surveillance face images are affected by many factors including low resolution, blur and noise. Therefore, how to restore a high quality face from a low quality face is a challenge. Face restoration technique provides a viable way to improve performance of facial analysis techniques on low quality face images.

Since face restoration has great potential in real applications, numerous face restoration algorithms have been proposed in recent years. Some algorithms focus on solving face restoration from low-resolution (LR) problem, such as works in \cite{wang2005hallucinating,ma2010hallucinating,wang2014face,yu2016ultra}. Other algorithms also take the noise corruption into consideration during face super resolution, such as works in \cite{jiang2014noise,jiang2016noise}. We observe that most existing face restoration methods omit one vital characteristic of real-world images, namely images in real applications always contain faces of different scales. Also, when the images are corrupted with serious distortions, it's hard to extract the faces in the distorted images for face restoration since face detection methods may not work well under this situation. Therefore, in this paper, we focus on solving face restoration close to real-world situation as illustrated in Figure \ref{fig:fig3}. Our proposed model can effectively restore face images with details from noise corrupted LR face images without scale limitation.

Face restoration can be considered as an image-to-image translation problem that transfers one image domain to another image domain. Solutions on image-to-image translation problem \cite{taigman2016unsupervised,yoo2016pixel,johnson2016perceptual} usually use autoencoder network \cite{hinton2006reducing} as a generator. However, single autoencoder network is too simple to represent multi-scale image-to-image translation due to lack of multi-scale representation. Meanwhile, ensemble learning, a machine learning paradigm where multiple learners are trained to solve the same problem, have shown its ability to make accurate prediction from multiple ``weak learners'' in classification problem \cite{Dietterich2000,kuncheva2004combining,polikar2006ensemble}. Therefore, an effective way to reinforce predictive performance of autoencoder network can be aggregating multiple base-generators into an enhanced-generator. In our model, we introduce base-encoders and base-decoders from low level to high level. These multi-level base-en/decoders ensure the generator have more diverse representation capacity to deal with multi-scale face image restoration.

The typical way of ensemble is to take a vote for the outputs of base-en/decoders. However, multi-scale face restoration is a problem that concerns multiple processes of feature extracting and restoring, merely taking a vote (or with other ensemble method) fails to incorporate high level information and restore detail information. Based on this observation, we devise a sequential ensemble structure that takes base-en/decoders from different levels as sequential data. The different combination directions of base-en/decoders are determined by the different goals of encoder and decoder. This sequential ensemble method is inspired by long short-term memory (LSTM) \cite{hochreiter1997long}. LSTM has been proved successfully in modeling sequential data, such as text and speech \cite{sundermeyer2012lstm,sutskever2014sequence}. The LSTM has the ability to optionally choose information passing through because of the gate mechanism. Specially, we design a Sequential Gating Unit (SGU) to realize information combination and selection, which sequentially takes base-en/decoders' information from two different levels as inputs and decides the output based on one active input.

Traditional optimization target of image restoration problem is to minimize the mean square error (MSE) between the restored image and the ground truth. However, minimizing MSE will often encourage smoothness in the restored image. Recently, generative adversarial networks (GANs) \cite{goodfellow2014generative,denton2015deep,radford2015unsupervised,salimans2016improved} show state-of-the-art performance on generating pictures of both high resolution and good semantic meaning. The high level real/fake decision made by discriminator causes the generated images looking real to the class of target domain. Therefore, we utilize the adversarial learning process proposed in GAN \cite{goodfellow2014generative} for restoration model training.

In general, we propose to solve multi-scale face restoration problem with a Sequential Gating Ensemble Network (SGEN). The contribution of our approach includes three aspects:
\begin{itemize}
  \item We employ the principle of ensemble learning into network architecture design. The SGEN is composed of multi-level base-en/decoders, which has better representation ability than ordinary autoencoder.
  \item The SGEN takes base-en/decoders from different levels as sequential data with two different directions corresponding to the different goals of encoder and decoder, which enables network to learn more compact high level information and restore more low level details.
  \item Furthermore, we propose a SGU unit to sequentially guide the combination of information from different levels.
\end{itemize}
The rest of this paper is organized as follows. We introduce related work in Section \ref{sec2} and present the details of proposed SGEN in Section \ref{sec3}, including network architecture, SGU unit and adversarial learning for SGEN. We present experiment results in Section \ref{sec4} and conclude in Section \ref{sec5}.
\section{Related Work}\label{sec2}
\begin{figure*}[!htpb]
  \centerline{\includegraphics[width=15.5cm]{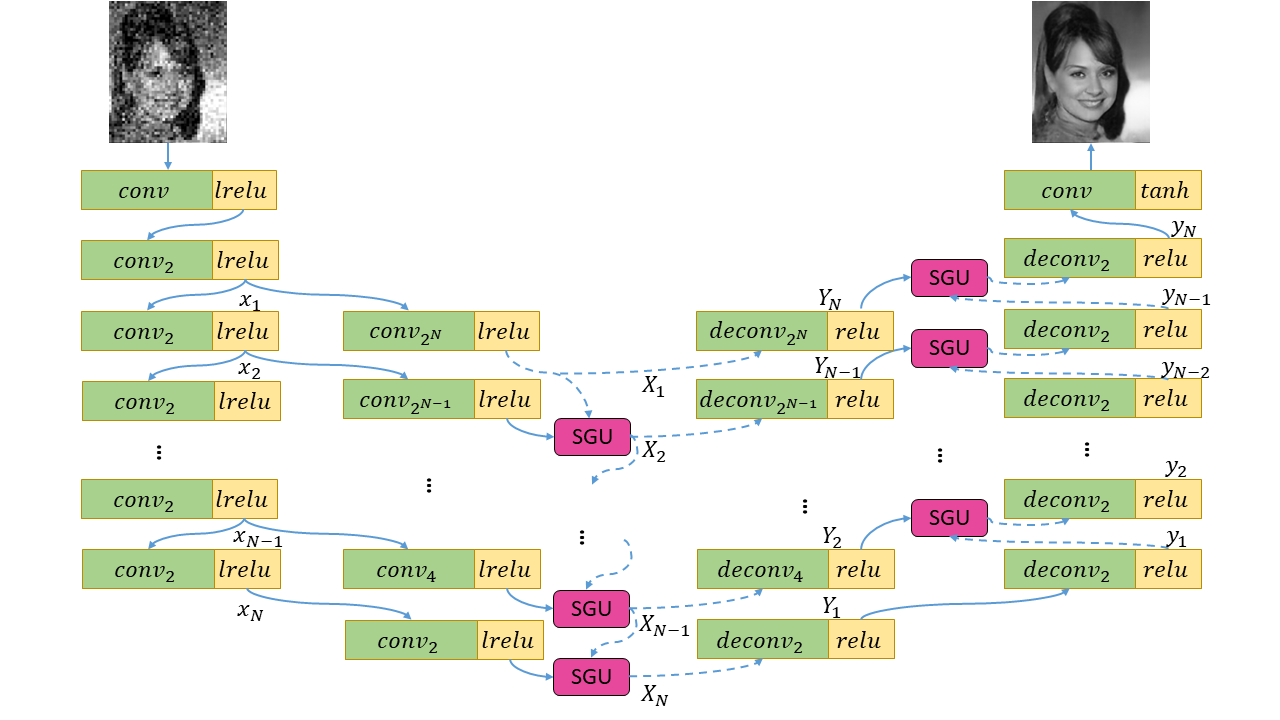}}
  \caption{Sequential ensemble network architecture of SGEN. Convolution and pooling operations are shown in green, activation functions are shown in yellow and the SGU is shown in pink.}
  \centering
\label{fig:fig1}
\end{figure*}
Face restoration has been studied for many years. In the early time, \cite{wang2005hallucinating} utilized global face-based method, i.e. principal component analysis (PCA), for face super-resolution. The work in \cite{ma2010hallucinating} proposed a least squares representation (LSR) framework that restores images using all the training patches, which incorporates more face priors. Due to the instability of LSR, \cite{wang2014face} introduced a weighted sparse representation (SR) with sparsity constraint for face super-resolution. However, one main drawback of SR based methods is its sensitivity to noise. Accordingly, \cite{jiang2014noise,jiang2016noise} introduced to reconstruct noise corrupted LR images with weighted local patch, namely locality-constrained representation (LcR).

Recently, convolutional neural networks (CNNs) based approaches have shown their superior performance in image restoration tasks. SRCNN \cite{dong2014learning} is a three layer fully convolutional network and trained end-to-end for image super resolution. \cite{yu2016ultra} presented a ultra-resolution discriminative generative network (URDGN) that can ultra-resolve a very low resolution face. Instead of building network as simple hierarchy structure, other works also applied the skip connections, which can be viewed as one kind of ensemble structure \cite{veit2016residual}, to image restoration tasks. \cite{ledig2016photo} proposed a SRResNet that uses ResNet blocks in the generative model and achieves state-of-the-art peak signal-to-noise ratio (PSNR) performance for image super-resolution. In addition, they presented a SRGAN that utilizes adversarial training to achieve better visual quality than SRResNet. \cite{mao2016image} proposed a residual encoder-decoder network (RED-Net) which symmetrically links convolutional and deconvolutional layers with skip-layer connections.

However, these skip-connections in \cite{ledig2016photo,mao2016image} fail to explore the underlying sequential relationship among multi-level feature maps in image restoration problem. Therefore, we design our SGEN followed by the goal of autoencoder, which sequentially extracts high level information from base-encoders in bottom-up manner and restores low level information from base-decoders in top-down manner.
\section{Sequential Gating Ensemble Network}\label{sec3}
Architecture of our Sequential Gating Ensemble Network (SGEN) is shown in Figure \ref{fig:fig1}. Details are discussed in the following subsections. We first introduce the sequential ensemble network architecture of SGEN. Then we present the Sequential Gating Unit (SGU) combining the multi-level information. Finally, we elaborate the adversarial training for SGEN and the loss function for adversarial training process.

\subsection{Sequential ensemble network architecture}
First, our generator is a fully convolutional computation network \cite{long2015fully} that can take arbitrary-size inputs and predict dense outputs. Then, let us denote $k$-$th$ encoder feature, $k$-$th$ base-encoder feature, $k$-$th$ combined base-encoder feature, $k$-$th$ base-decoder feature,  $k$-$th$ combined base-decoder feature by $x_{k}$, $X_{k}$, $\hat{X}_k$, $Y_{k}$, $\hat{Y}_k$ respectively, and suppose there are $N$ base-encoders and base-decoders in total. Given a low quality face image sample $s$, the SGEN $G$ in Figure \ref{fig:fig1} can be illustrated in the formulas below:
\begin{equation}\label{7}
{x_1} = lrelu(con{v_2}(lrelu(conv_1(s)))),
\end{equation}
\begin{equation}\label{8}
{x_k} = lrelu(con{v_2}({x_{k - 1}})), \quad k = 2,3,...,N
\end{equation}
\begin{equation}\label{9}
{X_k} = lrelu(con{v_{{2^{N-k+1}}}}({x_k})), \quad k = 1,2,...,N
\end{equation}
\begin{equation}\label{10-1}
{\hat{X}_1} = X_1,
\end{equation}
\begin{equation}\label{10}
{\hat{X}_k} = SGU(X_k, \hat{X}_{k-1}), \quad k = 2,3,...,N
\end{equation}

\begin{equation}\label{11}
{Y_k} = relu(decon{v_{2^{k}}}(\hat{X}_{N-k+1})), \quad k = 1,2,3,...,N
\end{equation}
\begin{equation}\label{12}
{\hat{Y}_1} = relu(decon{v_{2}}(Y_1)),
\end{equation}
\begin{equation}\label{13}
{\hat{Y}_k} = relu(decon{v_{2}}(SGU(Y_k, \hat{Y}_{k-1})), \quad k = 2,3,...,N
\end{equation}
\begin{equation}\label{14}
G(s) = tanh(conv_1(\hat{Y}_N)),
\end{equation}
where $G(s)$ is the generated face image, $con{v_{2^k}}$ and $decon{v_{2^k}}$ are convolution and de-convolution operations with factor $2^k$ pooling and upsampling respectively. SGU is sequential gating unit. Each de-convolution layer is followed by $relu$ (rectified linear unit) \cite{nair2010rectified}, and each convolution layer is followed by $lrelu$ (leaky relu) \cite{maas2013rectifier}, except for the last layer of generator (using $tanh$ activation function). Note, there is no parameters sharing among different convolution operations, de-convolution operations and SGUs.

The bottom-up base-encoders combination and top-down base-decoders combination are determined by the different goals of encoder and decoder. Given a low quality face image input, the encoder of a autoencoder would like to transfer the input into highly compact representation with semantic meaning (i.e., bottom-up information extraction), and the decoder would like to restore the face image with abundant details (i.e., top-down information restoration). Therefore, without breaking the rules of autoencoder, we combine the multi-level base-en/decoders in two directions. Accordingly, we design a SGU to realize the multi-level information combination and selection in en/decoder stage.

Combination of these multi-level base-en/decoders provides another benefit that network layer of SGEN contains multiple scales of receptive field, which helps the encoder learn features with multi-scale information and helps decoder generate more accurate images from multi-scale features. Experiment results also demonstrate that our network is more capable of restoring multi-scale low quality face image than other networks.
\subsection{Sequential gating unit}
\begin{figure}[!htpb]
  \centerline{\includegraphics[width=8.5cm]{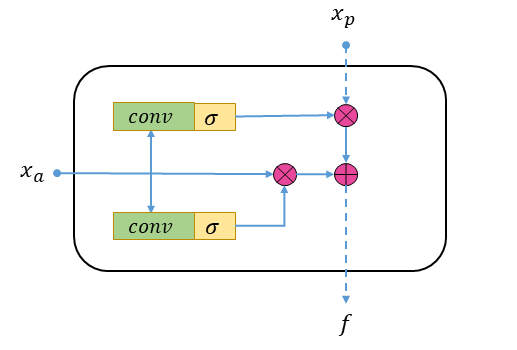}}
  \caption{Sequential Gating Unit. Element-wise multiplications and additions are shown in pink.}
  \centering
\label{fig:fig2}
\end{figure}
To further utilize the sequential relationship among multi-level base-en/decoders, we propose a Sequential Gating Unit (SGU) to sequentially combine and select the multi-level information, which takes base-en/decoders' information from two different levels as inputs and deciding the output based on one active input. The SGU is shown in the Figure \ref{fig:fig2}, equation depicting the unit is given as below:
\begin{equation}\label{15}
{f} = \sigma(conv(x_a))*x_a+\sigma(conv(x_a))*x_p,
\end{equation}
where $f$ is the SGU output, $\sigma(x)$ is sigmoid activation function representing ``gate'' in SGU, $x_a$ and $x_p$ are active input and passive input respectively. The active input $x_a$ makes the decision what information we are going to throw away from the passive input $x_p$ and what new information we are going to add from the active input itself. In the encoder stage, high level base-encoder acts as $x_a$ and takes control over the low level information, which sequentially updates the high level semantic information and removes noise. For the decoder stage, the low level base-decoder becomes $x_a$ and takes control over the high level information in an opposite direction, which sequentially restores low level information and generates images with more details.
\subsection{Adversarial training and loss function}
We apply the adversarial training of GAN in our proposed model. The adversarial training needs to learn a discriminator $D$ to guide the generator $G$, i.e. SGEN in this paper, to produce realistic target under the supervision of real/fake. In face restoration case, the objective function of GAN can be represented as minimax function:
\begin{equation}\label{16}
\begin{split}
\mathop {\min }\limits_G \mathop {\max }\limits_D L(D,G) &= {E_{t \sim {p_T}(t)}}[\log (D(t))]\\ &+{E_{s \sim {p_S}(s)}}[\log (1 - D(G(s))],
\end{split}
\end{equation}
where $s$ is a sample from the low quality source domain $S$ and $t$ is the corresponding sample in high quality target domain $T$. In addition to using adversarial loss in the generator training process, we add the mean square error (MSE) loss for generator to require the generated image $G(s)$ as close as to the ground truth value of pixels. The modified loss function for adversarial SGEN training is as below:
\begin{equation}\label{17}
\begin{split}
\mathop {\min }\limits_G \mathop {\max }\limits_D L(D,G) &= {E_{t \sim {p_T}(t)}}[\log (D(t))]\\&+{E_{s \sim {p_S}(s)}}[\log (1 - D(G(s))]\\&+\lambda {L_{MSE}}(G),
\end{split}
\end{equation}
\begin{equation}\label{18}
{L_{MSE}}(G)={E_{s \sim {p_S}(s),t \sim {p_T}(t)}}{[||t - G(s)|{|_2^2}]},
\end{equation}
where $\lambda$ is weight to achieve balance between adversarial term and MSE term.

To make the discriminator be able to take input of arbitrary size as well, we design a fully convolutional discriminator with global average pooling proposed in \cite{lin2013network}. We replace the traditional fully connected layer with global average pooling. The idea of global average pooling is to take the average of each feature map as the resulting vector fed into classification layer. Therefore, the discriminator has much fewer network parameters than fully connected network and overfitting is more likely to be avoided.

\section{Experiments}\label{sec4}
\subsection{Parameters setting}
In the experiments, we set $N=3$ levels for the SGEN to achieve a trade-off between performance and computation cost and the weight $\lambda$ is set to $0.1$. We use the adaptive learning method Adam \cite{kingma2014adam} as the optimization algorithm with learning rate of $0.0002$. Minibatch size is set to $64$ for every experiments.
\subsection{Dataset and evaluation metrics}
We carry out experiments below on the widely used face dataset CelebA \cite{liu2015faceattributes} containing $202599$ cropped celebrity faces. We set aside $30000$ images as test set, $20000$ images as validation set, and the rest as training set. We resize the face images from $128 \times 96$ to $208 \times 176$ which are commonly used resolutions in practice. Specially, we sample $6$ different scales between two resolutions. Then we downsample the images to low-resolution (LR) by a factor of $4$ and corrupt the images by AWGN noise (standard deviation $\sigma = 30$). To enable the noise corrupted LR images as the network input, we up-sample the LR images by a factor $4$ using nearest interpolation.

Restoration results are evaluated with peak signal-to-noise ratio (PSNR) and structure similarity index (SSIM) \cite{wang2004image}. However, PSNR and SSIM are objective metrics that can not always be consistent with human perceptual quality. Therefore, we conduct subjective evaluation for restoration results to further investigate the effectiveness of our model. We recruited $18$ subjects for subjective evaluation. We show each subject the corresponding source images prior to their evaluation for the generated images. Thus, they can form a general quality standard of generated images. After viewing one test image, subject gives a quality score from $1$ (bad quality) to $5$ (excellent quality). For each model, $288$ restoration images from six different scales are evaluated, and mean opinion scores (MOS) are computed at each scale.

\subsection{Comparison of different losses and ensemble methods}
\begin{figure}
  \centerline{\includegraphics[width=8.5cm]{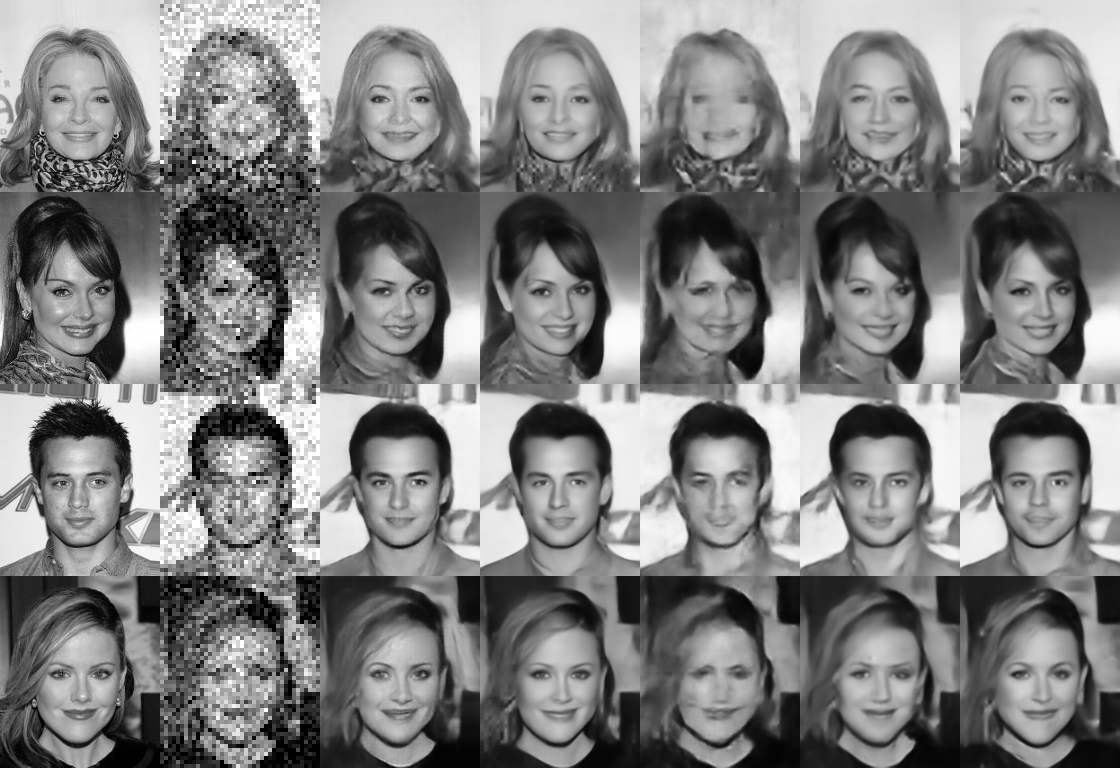}}
  \caption{Restoration samples from the same scale. From left to right: ground truth, noise corrupted LR images, results from SGEN, results from SGEN-MSE, results from MEN, results from AEN, results from CEN.}
  \centering
\label{fig:fig5}
\end{figure}
\begin{figure}
  \centerline{\includegraphics[width=8.5cm]{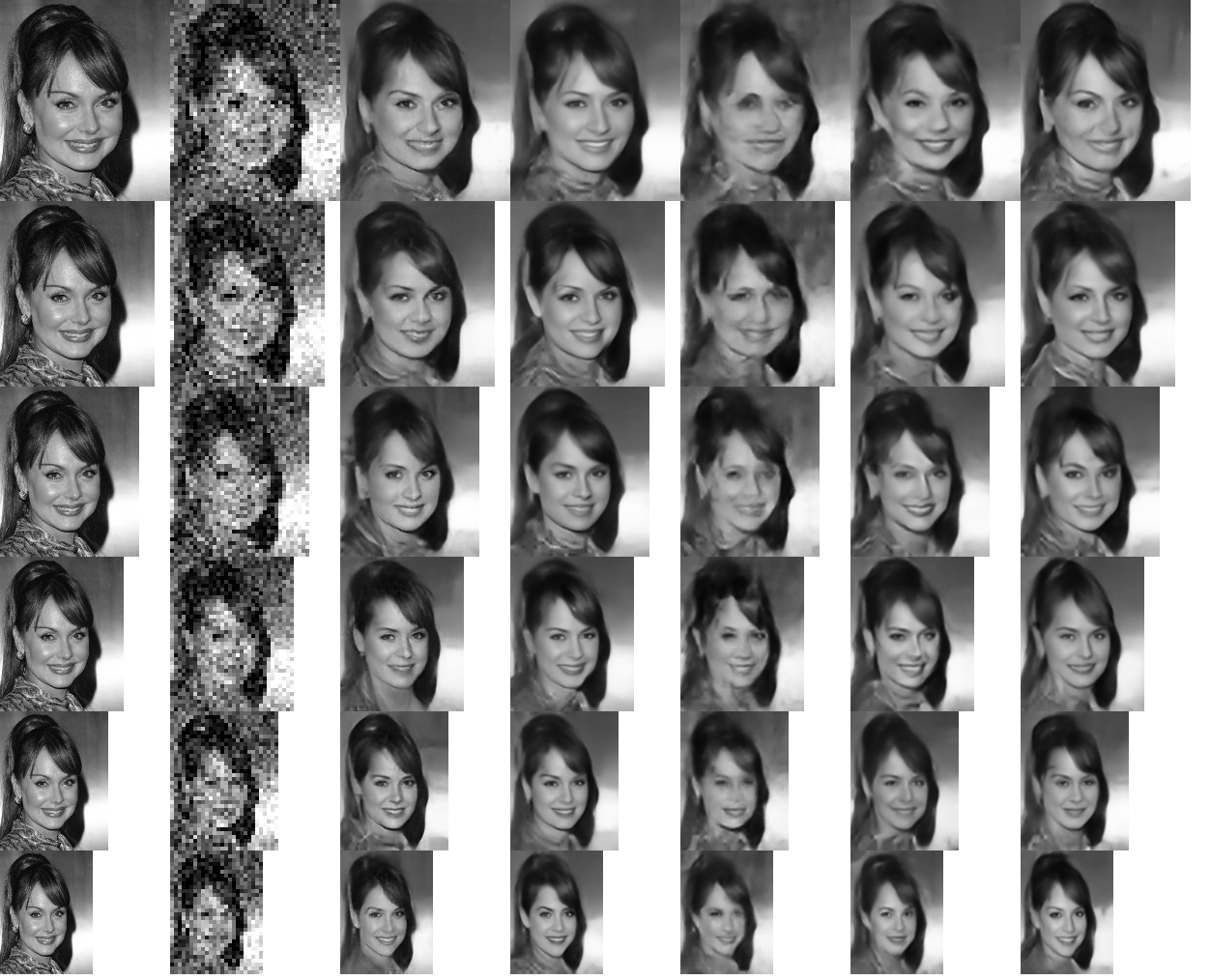}}
  \caption{Restoration samples from the six different scales. From left to right: ground truth, noise corrupted LR images, results from SGEN, results from SGEN-MSE, results from MEN, results from AEN, results from CEN.}
  \centering
\label{fig:fig7}
\end{figure}

\begin{table*}[]
\centering
\caption{Average PSNR, SSIM and MOS results of networks with different loss and ensemble methods from scale $128 \times 96$ to scale $160 \times 128$. Highest scores are in bold.}
\label{table3}
\begin{tabular}{|c|c|c|c|c|c|c|c|c|c|}
\hline
                  & \multicolumn{3}{c|}{Scale $128 \times 96$}                    & \multicolumn{3}{c|}{Scale $144 \times 112$}                   & \multicolumn{3}{c|}{Scale $160 \times 128$}                     \\ \hline
                  & \textbf{PSNR}  & \textbf{SSIM}   & \textbf{MOS} & \textbf{PSNR}  & \textbf{SSIM}  & \textbf{MOS} & \textbf{PSNR}  & \textbf{SSIM}   & \textbf{MOS} \\ \hline
\textbf{SGEN}     & 22.37         & 0.6555          &\textbf{4.4833}              & 23.08         & 0.6863          &\textbf{4.7833}              & 23.61          & 0.7006          &\textbf{4.6333}
              \\ \hline
\textbf{SGEN-MSE} & \textbf{23.00}   & \textbf{0.6989} &4.3667    & \textbf{23.60} & \textbf{0.7161} &4.5000    & \textbf{24.12} & \textbf{0.7327} &4.6000              \\ \hline
\textbf{MEN}      & 22.10          & 0.6485          &2.0833              & 22.68         & 0.6652          &2.1333              & 23.04          & 0.6807          &2.0833              \\ \hline
\textbf{AEN}      & 22.61         & 0.6800            &3.2000              & 23.17         & 0.7015          &3.4000              & 23.64          & 0.7134          &3.2000              \\ \hline
\textbf{CEN}      & 22.75         & 0.6945          &4.0333              & 23.36         & 0.7129          &4.1000              & 23.88          & 0.7261          &4.2000              \\ \hline
\end{tabular}
\end{table*}
\begin{table*}[]
\centering
\caption{Average PSNR, SSIM and MOS results of networks with different loss and ensemble methods from scale $176 \times 144$ to scale $208 \times 176$. Highest scores are in bold.}
\label{table4}
\begin{tabular}{|c|c|c|c|c|c|c|c|c|c|}
\hline
                  & \multicolumn{3}{c|}{Scale $176 \times 144$}                    & \multicolumn{3}{c|}{Scale $192 \times 160$}                   & \multicolumn{3}{c|}{Scale $208 \times 176$}                   \\ \hline
                  & \textbf{PSNR}  & \textbf{SSIM}   & \textbf{MOS} & \textbf{PSNR}  & \textbf{SSIM}  & \textbf{MOS} & \textbf{PSNR} & \textbf{SSIM}   & \textbf{MOS} \\ \hline
\textbf{SGEN}     & 24.12          & 0.7202          &\textbf{4.6333}              & 24.55          & 0.733           &4.6000              & 24.92          & 0.7429          &\textbf{4.6833}              \\ \hline
\textbf{SGEN-MSE} & \textbf{24.61} & \textbf{0.7501} &4.3667    & \textbf{24.98} & \textbf{0.7576} & \textbf{4.6333}    & \textbf{25.39} & \textbf{0.7686} &4.5833              \\ \hline
\textbf{MEN}      & 23.50           & 0.6942          &1.9833              & 23.97          & 0.7052          &2.1167              & 24.46          & 0.7197          &2.2667              \\ \hline
\textbf{AEN}      & 24.14          & 0.7274          &3.2333              & 24.62          & 0.7411          &3.3000              & 24.97          & 0.7533          &3.2667              \\ \hline
\textbf{CEN}      & 24.42          & 0.7397          &4.1000              & 24.82          & 0.7545          &4.0500              & 25.20           & 0.7646          &4.0833              \\ \hline
\end{tabular}
\end{table*}

To investigate influence of different loss choices and verify the effectiveness of sequential ensemble structure of SGEN, we compare performance of following models:
\begin{itemize}
  \item \textbf{SGEN}. SGEN is trained with MSE and adversarial loss.
  \item \textbf{SGEN-MSE}. SGEN-MSE is trained with only MSE loss.
  \item \textbf{MEN}. MEN (Max Ensemble Network) is just like SGEN but without any SGU during the combination of base-en/decoders, it uses max ensemble instead.
  \item \textbf{AEN}. AEN (Average Ensemble Network) uses average ensemble instead of SGU.
  \item \textbf{CEN}. AEN (Concatenate Ensemble Network) uses concatenate ensemble instead of SGU.
\end{itemize}
The objective and subjective results are shown in Table \ref{table3} and Table \ref{table4}. Visual restoration samples from one scale and six scales are shown in Figure \ref{fig:fig5} and Figure \ref{fig:fig7}. Only MSE loss for SGEN training provides higher PSNR and SSIM than combining MSE and adversarial loss, this result is not surprising because only minimizing MSE is equal to maximize PSNR. Smoother restoration results are also obtained by SGEN-MSE and are less visually preferred than SGEN through MOS evaluation. Comparing SGEN with other non-sequential ensemble networks, SGEN achieves better subjective scores and produces visually preferred images with more face details. Actually, because of the gate mechanism in SGU, sequential ensemble method can be viewed as a generalized ensemble method including max ensemble and average ensemble. The CEN shows more competitive results than MEN and AEN because CEN combines base-en/decoders' information all together without any information selection. The better performance of SGEN than CEN also demonstrates the effectiveness of automatically information selection in sequential ensemble method.

\subsection{Comparison with state-of-the-art algorithms}\label{4.4}
\begin{table*}[]
\centering
\caption{Average PSNR, SSIM and MOS comparison results with state-of-the-art algorithms from scale $128 \times 96$ to scale $160 \times 128$. Highest scores are in bold.}
\label{table1}
\begin{tabular}{|c|c|c|c|c|c|c|c|c|c|}
\hline
                  & \multicolumn{3}{c|}{Scale $128 \times 96$}                    & \multicolumn{3}{c|}{Scale $144 \times 112$}                   & \multicolumn{3}{c|}{Scale $160 \times 128$}                     \\ \hline
                  & \textbf{PSNR}  & \textbf{SSIM}   & \textbf{MOS} & \textbf{PSNR}  & \textbf{SSIM}  & \textbf{MOS} & \textbf{PSNR}  & \textbf{SSIM}   & \textbf{MOS} \\ \hline
\textbf{SGEN}     & 22.37         & 0.6555          &\textbf{4.4833}              & 23.08         & 0.6863          &\textbf{4.7833}              & 23.61          & 0.7006          &\textbf{4.6333}
              \\ \hline
\textbf{SGEN-MSE} & \textbf{23.00}   & \textbf{0.6989} &4.3667    & \textbf{23.60} & \textbf{0.7161} &4.5000    & \textbf{24.12} & \textbf{0.7327} &4.6000              \\ \hline
SRCNN             & 21.72         & 0.5923          &1.0000              & 22.22         & 0.6094          &1.0667              & 22.69          & 0.6236          &1.1000              \\ \hline
SRResNet          & 22.73         & 0.6827          &3.0333              & 23.29         & 0.7016          &3.0667              & 23.81          & 0.7166          &3.2667              \\ \hline
SRGAN          & 22.29         & 0.6486          &2.9444              & 22.78         & 0.6796          &3.0000             & 23.43          & 0.6927          &3.3889              \\ \hline
RED-Net           & 22.77         & 0.6809          &3.6667              & 23.32         & 0.7001          &3.6333              & 23.83          & 0.7147          &3.8167              \\ \hline
URDGN             & 22.54         & 0.6688          &2.8667              & 23.10          & 0.6885          &3.0667              & 23.56          & 0.7044          &3.0500              \\ \hline
\end{tabular}
\end{table*}
\begin{table*}[]
\centering
\caption{Average PSNR, SSIM and MOS comparison results with state-of-the-art algorithms from scale $176 \times 144$ to scale $208 \times 176$. Highest scores are in bold.}
\label{table2}
\begin{tabular}{|c|c|c|c|c|c|c|c|c|c|}
\hline
                  & \multicolumn{3}{c|}{Scale $176 \times 144$}                    & \multicolumn{3}{c|}{Scale $192 \times 160$}                   & \multicolumn{3}{c|}{Scale $208 \times 176$}                   \\ \hline
                  & \textbf{PSNR}  & \textbf{SSIM}   & \textbf{MOS} & \textbf{PSNR}  & \textbf{SSIM}  & \textbf{MOS} & \textbf{PSNR} & \textbf{SSIM}   & \textbf{MOS} \\ \hline
\textbf{SGEN}     & 24.12          & 0.7202          &\textbf{4.6333}              & 24.55          & 0.7330           &4.6000              & 24.92          & 0.7429          &\textbf{4.6833}              \\ \hline
\textbf{SGEN-MSE} & \textbf{24.61} & \textbf{0.7501} &4.3667    & \textbf{24.98} & \textbf{0.7576} & \textbf{4.6333}    & \textbf{25.39} & \textbf{0.7686} &4.5833              \\ \hline
SRCNN             & 23.17          & 0.6419          &1.1500              & 23.58          & 0.6548          &1.2333              & 23.92          & 0.6643          &1.4833              \\ \hline
SRResNet          & 24.31          & 0.7328          &3.3500              & 24.77          & 0.7458          &3.0833              & 25.04          & 0.7543          &3.1500              \\ \hline
SRGAN          & 24.03          & 0.7198          &3.8333              & 24.37          & 0.7297          &3.9444              & 24.71          & 0.7413          &3.8889              \\ \hline
RED-Net           & 24.30           & 0.7318          &3.6667              & 24.73          & 0.7461          &3.6500              & 25.11          & 0.7560           &3.7167              \\ \hline
URDGN             & 23.98          & 0.7181          &2.9333              & 24.42          & 0.7306          &2.9500              & 24.67          & 0.7396          &2.8500              \\ \hline
\end{tabular}
\end{table*}
\begin{figure}
  \centerline{\includegraphics[width=8.5cm]{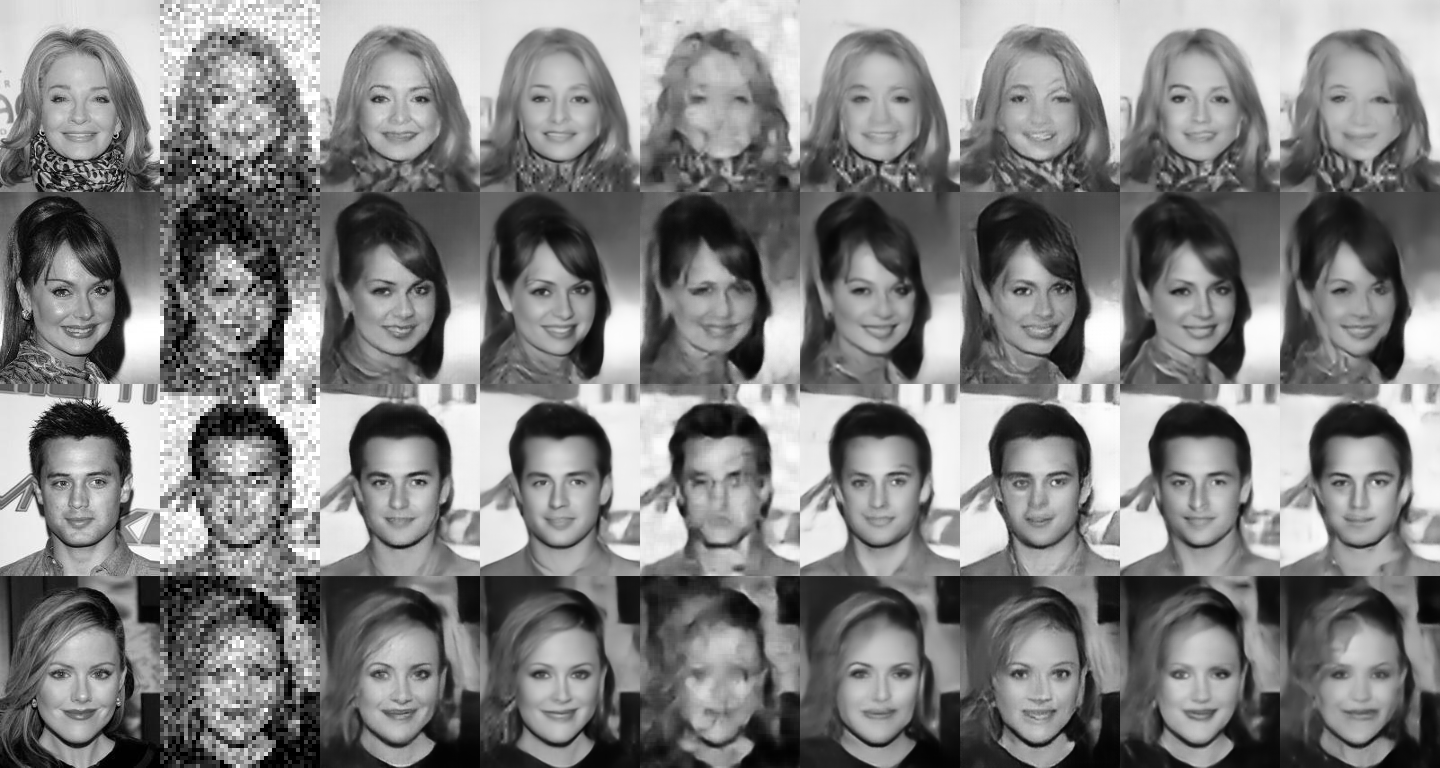}}
  \caption{Restoration samples from the same scale. From left to right: ground truth, noise corrupted LR images, results from SGEN, results from SGEN-MSE, results from SRCNN, results from SRResNet, results from SRGAN, results from RED-Net, results from URDGN.}
  \centering
\label{fig:fig6}
\end{figure}
We compare the performance of SGEN with five state-of-the-art image restoration networks: SRCNN \cite{dong2014learning}, SRResNet\cite{ledig2016photo}, SRGAN\cite{ledig2016photo}, RED-Net \cite{mao2016image} and URDGN \cite{yu2016ultra}. All the networks are retrained on the same multi-scale and noisy face dataset. The quantitative results are shown in Table \ref{table1} and Table \ref{table2}. The visual restoration samples from one single scale and six scales are shown in Figure \ref{fig:fig6} and Figure \ref{fig:fig8}. The quantitative results confirm that our SGEN with only MSE loss achieves state-of-the-art performance in terms of PSNR and SSIM. MOS results from subjective evaluation also suggest that SGEN with adversarial training produces better perceptual quality than other algorithms. From Figure \ref{fig:fig6} and Figure \ref{fig:fig8}, we can see that results from SGEN are more visually clear and contain more face details. Compared with non-ensemble structure SRCNN and URGAN, our SGEN shows better ability to handle multi-scale face restoration. Although the RED-Net, SRResNet and SRGAN try to restore image details with skip-connections that can be viewed as one kind of ensemble structure \cite{veit2016residual}, the networks can not restore faces as good as SGEN due to lack of sequential ensemble method that can sequentially choose to extract high level information from corrupted input and restore more low level details.
\subsection{Computational Time}\label{4.5}
We train all the networks on one NVIDIA K80 GPU. The training time of the SGEN (average training time = $26$ GPU-hours) is little slower than the structure with the same loss terms but without ensemble, such as URDGN (average training time = $20$ GPU-hours). The sequential ensemble structure converges little slower than the single autoencoder structure is not surprising, as the SGEN learns multi-scale features and possesses a larger feature capacity. Considering the fact that our network can converge slightly more than one GPU-day, which means that our network structure is a relatively light structure and the convergence is not a problem for our network. In addition, our model only takes about $0.016s$ on average for each test image, which is well suited for real-time image processing.
\subsection{Influence of $N$}\label{4.6}
\begin{table}[]
\centering
\caption{Average PSNR and SSIM performance and computational cost time of SGEN with different $N$.}
\label{table5}
\begin{tabular}{|c|c|c|c|}
\hline
              & \textit{\textbf{N=2}} & \textit{\textbf{N=3}} & \textit{\textbf{N=4}} \\ \hline
\textbf{PSNR} & 24.13        & 24.28        & 24.49        \\ \hline
\textbf{SSIM} & 0.7289       & 0.7373       & 0.7463       \\ \hline
\textbf{Time} & 20h          & 26h          & 35h          \\ \hline
\end{tabular}
\end{table}
In addition to SGEN with $N=3$, we train other two SGENs to explore the influence of main parameter $N$ of SGEN. The average PSNR and SSIM performance on test set, and computational cost time of SGEN with different $N$ are shown in Table \ref{table5}. Without losing the universality, only MSE loss for SGEN is used for this comparison. It is obvious that the performance of SGEN improves with the increase of $N$. SGEN with $N=4$ achieves performance gain about $0.86\%$ in terms of PSNR against SGEN with $N=3$. However, $35\%$ extra computational cost will also be added by SGEN with $N=4$ compared with $N=3$. Therefore, we choose $N=3$ in our paper to achieve a trade-off between computational cost and performance.
\begin{figure}
  \centerline{\includegraphics[width=8.5cm]{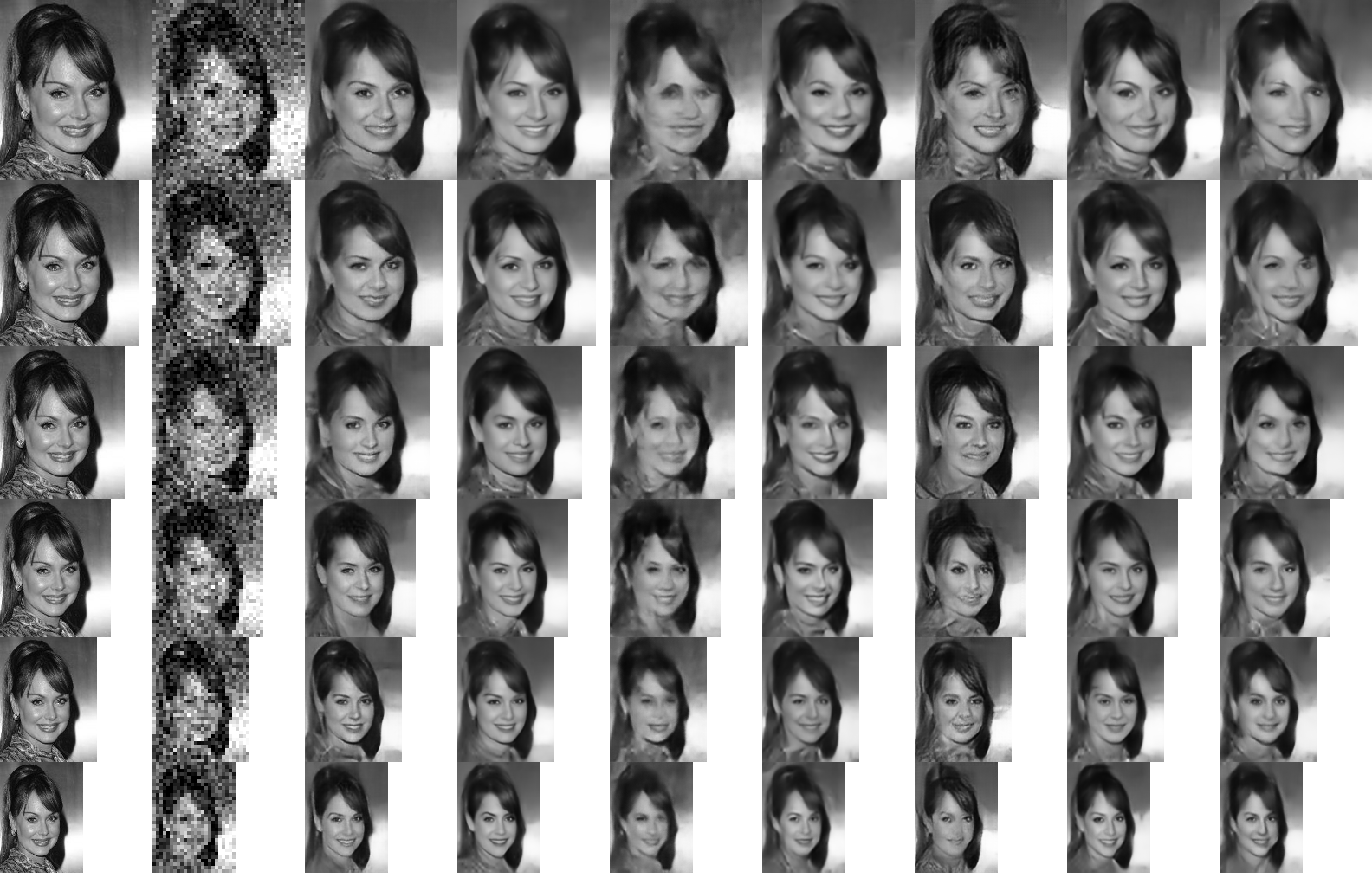}}
  \caption{Restoration samples from the six different scales. From left to right: ground truth, noise corrupted LR images, results from SGEN, results from SGEN-MSE, results from SRCNN, results from SRResNet, results from SRGAN, results from RED-Net, results from URDGN.}
  \centering
\label{fig:fig8}
\end{figure}
\section{Conclusions and Future Work}\label{sec5}
In this paper, we present a SGEN model for multi-scale face restoration from low-resolution and strong noise. We propose to aggregate multi-level base-en/decoders into the SGEN. The SGEN takes en/decoders from different levels as the sequential data. Specifically, the SGEN learns to sequentially extract high level information from low level base-encoders in bottom-up manner and sequentially restore low level details from high level base-decoders in top-down manner. Specially, we propose an elaborate SGU unit that could sequentially combine and select the multi-level information from base-en/decoders. Owing to the sequential ensemble structure, SGEN with MSE loss achieves the state-of-the-art performance of multi-scale face restoration in terms of PSNR and SSIM. Further applying adversarial loss to SGEN training, SGEN achieves the best perceptual quality according to subjective evaluation.

There are multiple aspects to explore for SGEN. First, we will apply the proposed model to other image-to-image translation tasks. Second, it is worth exploring SGEN on face video restoration in the future, which can be utilized for real-time surveillance video analysis. Third, it is also interesting to combine SGEN with face analysis techniques for end-to-end face restoration and face analysis.
\section{Acknowledgement}
This work was supported in part by the National Key Research and Development Program of China under Grant No. 2016YFC0801001, NSFC under Grant 61571413, 61632001,61390514, and Intel ICRI MNC.

\bibliography{Bibliography-File}
\bibliographystyle{aaai}
\end{document}